\pdfoutput=1

\documentclass[11pt]{article}

\usepackage[preprint]{acl}

\usepackage{times}
\usepackage{latexsym}

\usepackage[T1]{fontenc}
\usepackage{enumitem}
\usepackage[utf8]{inputenc}
\usepackage{multirow}
\usepackage{microtype}
\usepackage{bbding}
\usepackage{inconsolata}
\usepackage{amssymb}
\usepackage[table]{xcolor}
\usepackage{graphicx}

\usepackage{amsmath}
\usepackage{booktabs}
\usepackage{array}

\title{CrowdAgent: Multi-Agent Managed Multi-Source Annotation System}

\author{
    \textbf{Maosheng Qin\textsuperscript{1}\thanks{~~Equal contribution.}},
    \textbf{Renyu Zhu\textsuperscript{2}$^*$},
    \textbf{Mingxuan Xia\textsuperscript{1}},
    \textbf{Chenkai Chen\textsuperscript{3}},
    \textbf{Zhen Zhu\textsuperscript{3}},
    \textbf{Minmin Lin\textsuperscript{2}},
\\
    \textbf{Junbo Zhao\textsuperscript{1}},
    \textbf{Lu Xu\textsuperscript{2}},
    \textbf{Changjie Fan\textsuperscript{2}},
    \textbf{Runze Wu\textsuperscript{2} \thanks{~~Corresponding authors.}},
    \textbf{Haobo Wang\textsuperscript{1} $^\dag$}
\\
    \textsuperscript{1}Zhejiang University
    \textsuperscript{2}NetEase Fuxi AI Lab
    \textsuperscript{3}Zhejiang Sci-tech University \\
    \texttt{qin.mmms@gmail.com, wanghaobo@zju.edu.cn}\\
}

\begin{document}
\maketitle
\begin{abstract}

High-quality annotated data is a cornerstone of modern Natural Language Processing (NLP). While recent methods begin to leverage diverse annotation sources—including Large Language Models (LLMs), Small Language Models (SLMs), and human experts—they often focus narrowly on the labeling step itself. A critical gap remains in the holistic \textbf{process control} required to manage these sources dynamically, addressing complex scheduling and quality-cost trade-offs in a unified manner. Inspired by real-world crowdsourcing companies, we introduce \textit{CrowdAgent}, a multi-agent system that provides end-to-end process control by integrating task assignment, data annotation, and quality/cost management. It implements a novel methodology that rationally assigns tasks, enabling LLMs, SLMs, and human experts to advance synergistically in a collaborative annotation workflow. We demonstrate the effectiveness of CrowdAgent through extensive experiments on six diverse multimodal classification tasks. The source code and video demo are available at \href{https://github.com/qmmms/crowdagent}{\texttt{https://github.com/QMMMS/CrowdAgent}}. 

\end{abstract}

\section{Introduction}

High-quality annotated data serves as the cornerstone for advancements in machine learning models and is pivotal for the digital transformation of enterprises. 
To meet the substantial data requirements, manual annotation through crowdsourcing \citep{DBLP:conf/icde/LiWZF17,DBLP:journals/smr/ZhenKNZAK21,DBLP:journals/ieeejas/Zhang22} has been a prevalent approach due to its inherent scalability and flexibility. However, this method faces challenges, including high costs, particularly for tasks demanding specialized expertise, such as medical \citep{johnson2023mimic, 10.1007/s11280-022-01013-6} and financial \citep{chen-etal-2021-finqa} tasks, and is labor-intensive for large-scale annotation.

\begin{figure}[t]
  \includegraphics[width=\columnwidth]{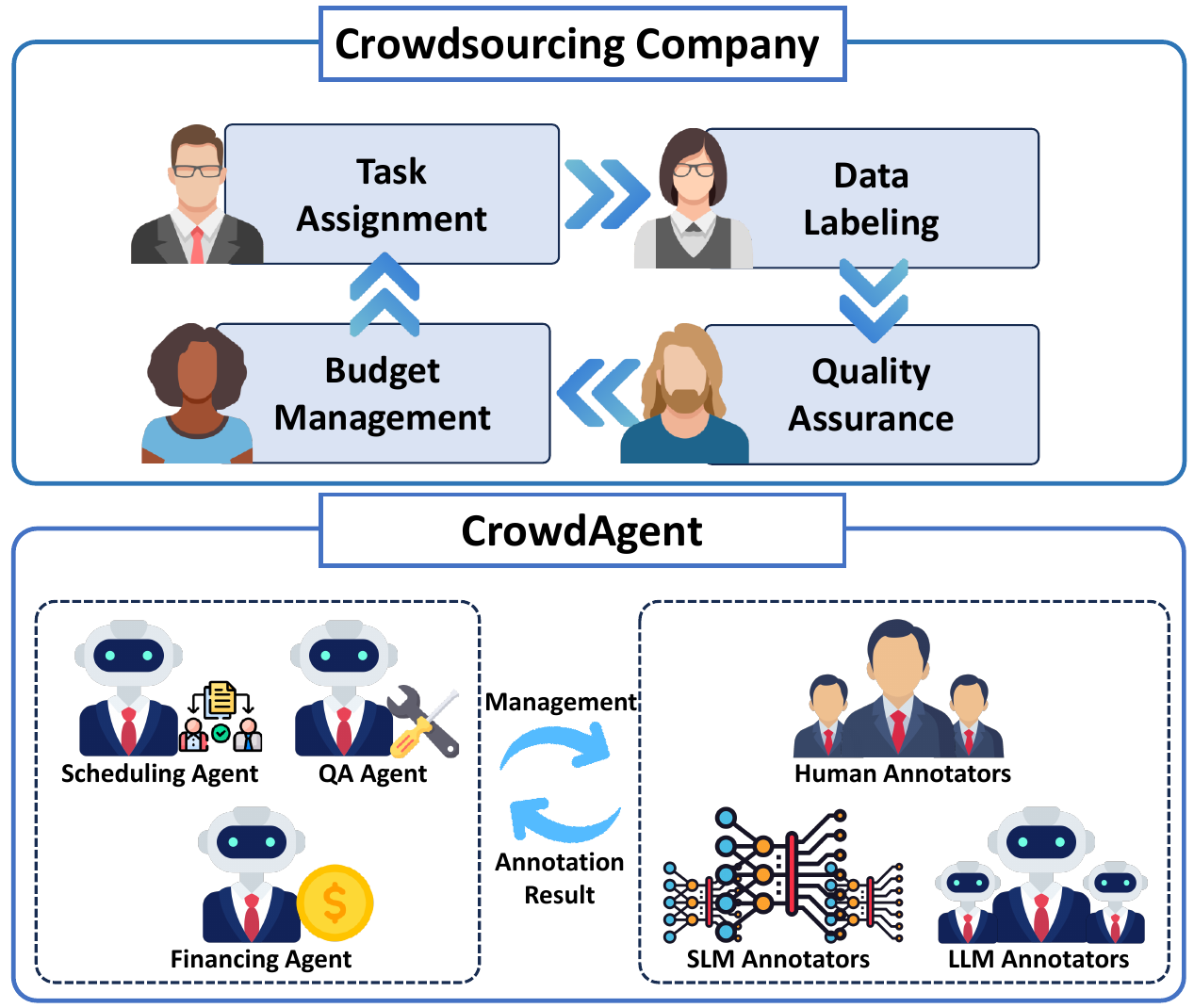}
  \caption{Comparison between traditional crowdsourcing company and our proposed system. CrowdAgent automates human management by using a collaborative multi-agent system, where tasks are dynamically dispatched to diverse annotation sources.}
  \label{fig:intro}
\end{figure}

To address this, LLMs-based annotation methods \citep{ding-etal-2023-gpt} have been particularly prevalent due to LLMs' strong zero-shot capabilities that achieve near-human performance while incurring lower costs. 
Due to inherent biases that can lead to inaccurate results, relying solely on LLMs is often insufficient. To address these shortcomings, recent work has focused on creating hybrid systems that combine multiple annotation sources. 
For example, Co-Annotating \cite{li-etal-2023-coannotating} treats LLMs as diverse annotators by employing a set of 7 prompt types; FreeAL \citep{xiao-etal-2023-freeal} is motivated by classical active learning theory to integrates noisy-label trained small models as a label filter; Lapras \citep{10.1145/3613904.3641960} utilizes verification scores to combine human and LLMs. 
However, current solutions mostly rely on a predefined labeling task with a fixed cost budget and seek to maximize label quality via algorithmic design. 

Moreover, a significant disconnect persists between algorithmic techniques and the practical \textbf{annotation process control} required in real-world projects. As depicted in Figure \ref{fig:intro}, in a human-run crowdsourcing company, when a user submits an annotation task with a specified budget, a \textit{manager} selects an appropriate group of\textit{ human annotators}. After the initial annotations are collected, \textit{quality assurance reviewers} assess their accuracy, and each annotator's profile is updated to inform future task assignment. A \textit{finance officer} then reviews the expenditures. Finally, the aggregated report on accuracy and cost is returned to the \textit{manager}, who decides whether to re-assign the task for another round of annotation or to deliver the final results.
Technically, to achieve such a \textit{complex, multi-stage process}, multi-agent system \citep{DBLP:conf/icml/HuangAPM22, lin2025decodingtimeseriesllms, 
DBLP:journals/fcsc/WangMFZYZCTCLZWW24, chen-si-2024-reflections} emerge as a natural and powerful solution to intelligently control the entire process, in order to assign tasks to the most suitable annotators under evolving conditions, ultimately maximizing annotation efficiency. Yet, such a holistic annotation agent system remains underexplored.

In this paper, we introduce \textbf{\textit{CrowdAgent}}, a novel multi-agent system to manage the dynamic collaboration between multiple annotation sources, with real-time quality assurance and cost management. 
Our system comprises four core components:
1) The \textbf{Annotation Agents} comprise multi-source annotators, including various large models, small models, and humans\footnote{We generalize human as an agent concept since it is indispensable to introduce human knowledge in annotation.}, providing labels for the given annotation task; 2) The \textbf{Quality Assurance (QA)  Agent} evaluates label quality using golden samples and performs label aggregation; 3) The \textbf{Financing Agent} monitors real-time budget consumption and evaluates the cost-effectiveness of each annotation source; 4) The \textbf{Scheduling Agent} dynamically dispatches tasks by synthesizing performance history, cost analysis, and quality feedback to rationally assign each sample to the most suitable annotator, thereby achieving an optimal trade-off between label quality and annotation cost.

Our CrowdAgent System also provides a user-friendly and extensible annotation platform. The core functionalities include flexible task configuration, real-time monitoring of agent interactions, and visualization of key metrics such as labeling accuracy and budget consumption. The system closely simulates the operational workflows of real-world crowdsourcing platforms, enabling users to dynamically adapt scheduling strategies based on runtime feedback. This powerful human-in-the-loop collaboration, designed to optimize the quality-cost trade-off, empowers enterprises and research institutions to build high-quality datasets more efficiently for the fast deployment of AI applications.

\begin{figure*}[t]
  \centering
  \includegraphics[width=2\columnwidth]{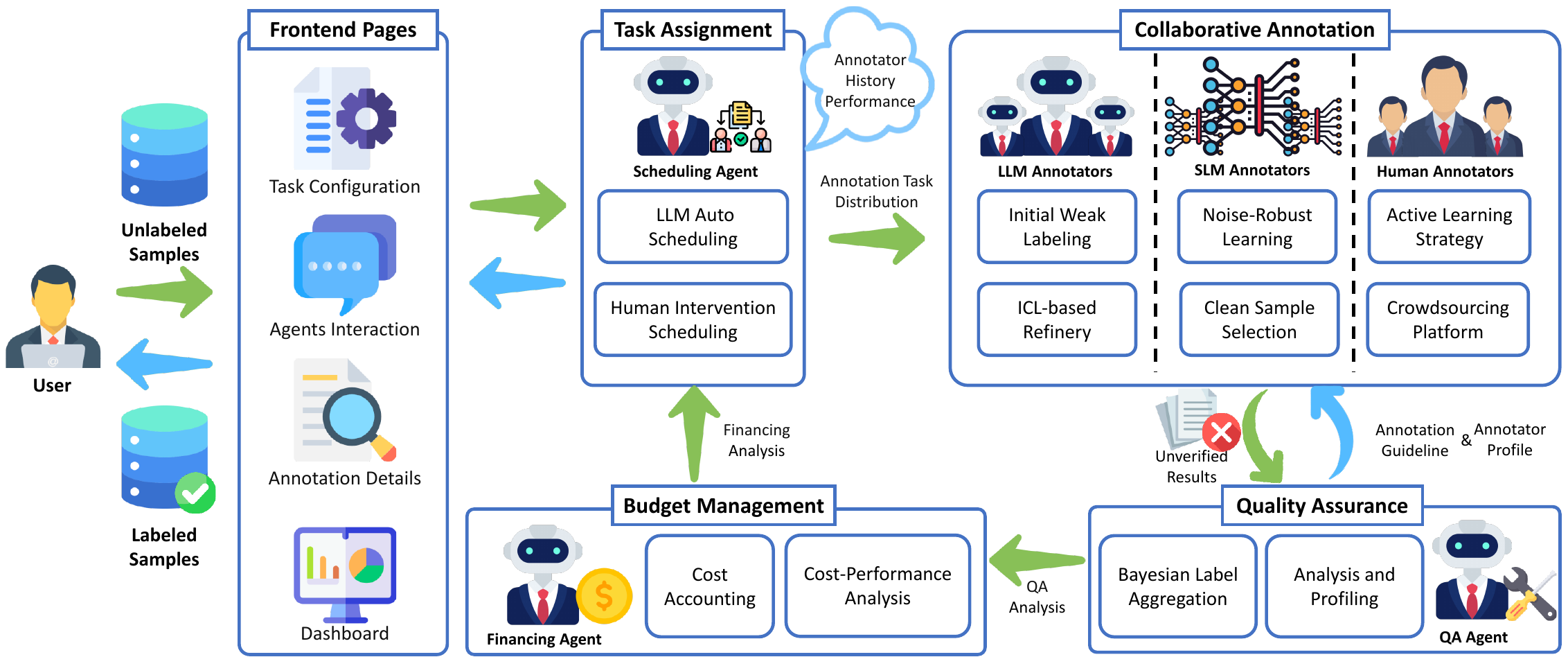}
  \caption{Workflow of CrowdAgent. The system emulates a virtual annotation company by employing intelligent agents with specialized skills. These agents orchestrate a collaborative workflow, dispatching tasks to a diverse set of annotation sources to leverage their respective strengths.}
  \label{fig:algo pic}
\end{figure*}

\section{The CrowdAgent System}

\paragraph{Notations.} Formally, the user initiates the process by defining an annotation task and providing an unlabeled dataset $\mathcal{D} = \{x_i\}_{i=1}^{N}$ with $N$ samples, where the corresponding true label $y_i \in \mathcal{Y} = \{1, \dots, C\}$ of $x_i$ is initially unknown. Here, $C$ represents the total number of classes. To evaluate annotation quality, a labeled golden set $\mathcal{D}_{\text{labeled}} = \{(x_i, y_i)\}_{i=1}^{M}$ is secretly mixed into the main dataset, which can either be curated from prior, similar annotation tasks or be formed by pre-labeling a small portion of $\mathcal{D}$. To prevent label leakage, golden set is used \textit{exclusively} by the Quality Assurance Agent for label aggregation and profiling. $\mathcal{D}_{\text{labeled}}$ is never exposed to the annotators or used as in-context learning examples in prompts.

\paragraph{An Overview.} 
CrowdAgent establishes a multi-agent annotation workflow (see Figure \ref{fig:algo pic}) that mirrors the roles and interactions of its human counterpart. Each agent within the system is defined by a specific role and a set of operational skills, and all agents adhere to the ReAct-style behavior described by \citet{yao2023react}. Communication between agents is facilitated through the exchange of structured files and a shared message pool. All agents in CrowdAgent are designed to learn from the decisions and outcomes of previous rounds.
\textit{
Due to the page limits, we refer the readers to Appendix \ref{sec:tech detail} for more technical details such as prompt designs and model training methodologies. 
}

\subsection{Annotation Agents}\label{sec:Annotation Agent}

Firstly, we describe Annotation Agents---our core operational units directly involved in the data labeling process. Traditional annotation techniques relying on a single source struggle to simultaneously optimize for efficiency, accuracy, and cost. For instance, manual annotation by human experts, while often high-quality, is expensive and lacks scalability. Large-scale annotation by LLMs is efficient but can be susceptible to biases and noise. To mitigate these individual drawbacks, our Annotation Agents consist of three main categories:

\noindent\textbf{(1) LLM Annotators:} Following \citet{li-etal-2023-coannotating}, our system instantiates multiple, distinct LLM annotator agents by employing several types of prompt designs. These prompts are engineered to guide each LLM annotator to adopt a different perspective on the task. The strategies include introducing a confirmation bias, altering the task structure via sequence swapping, and reframing the problem by changing the question format (e.g., True/False or Multiple Choice). This diversity in prompting and in-context learning using dynamically curated examples enhances robustness.

\noindent\textbf{(2) SLM Annotators:} Motivated by previous work \citep{xiao-etal-2023-freeal}, we train small deep models with noisy-label learning techniques \citep{li2020dividemix}. They serve as label purification agents that distill potential true labels from noisy LLM outputs. The detailed training process is shown in Appendix \ref{sec:collaborative-annotation}. 
    
\noindent\textbf{(3) Human Annotators:} Humans provide expert judgment on the most challenging samples, which are strategically selected based on low confidence and diversity metrics to maximize their impact. To seamlessly integrate it with our system, we consider an off-the-shelf online \textit{NetEase Youling Crowdsourcing Platform} \footnote{https://zb.163.com/} due to its native support for programmatic task dispatch. One may also consider other crowdsourcing platforms like \textit{Amazon Mechanical Turk} \cite{10.1007/978-3-642-35142-6_14}. 

\subsection{Quality Assurance Agent}

\begin{figure*}[t]
  \centering
  \includegraphics[width=2\columnwidth]{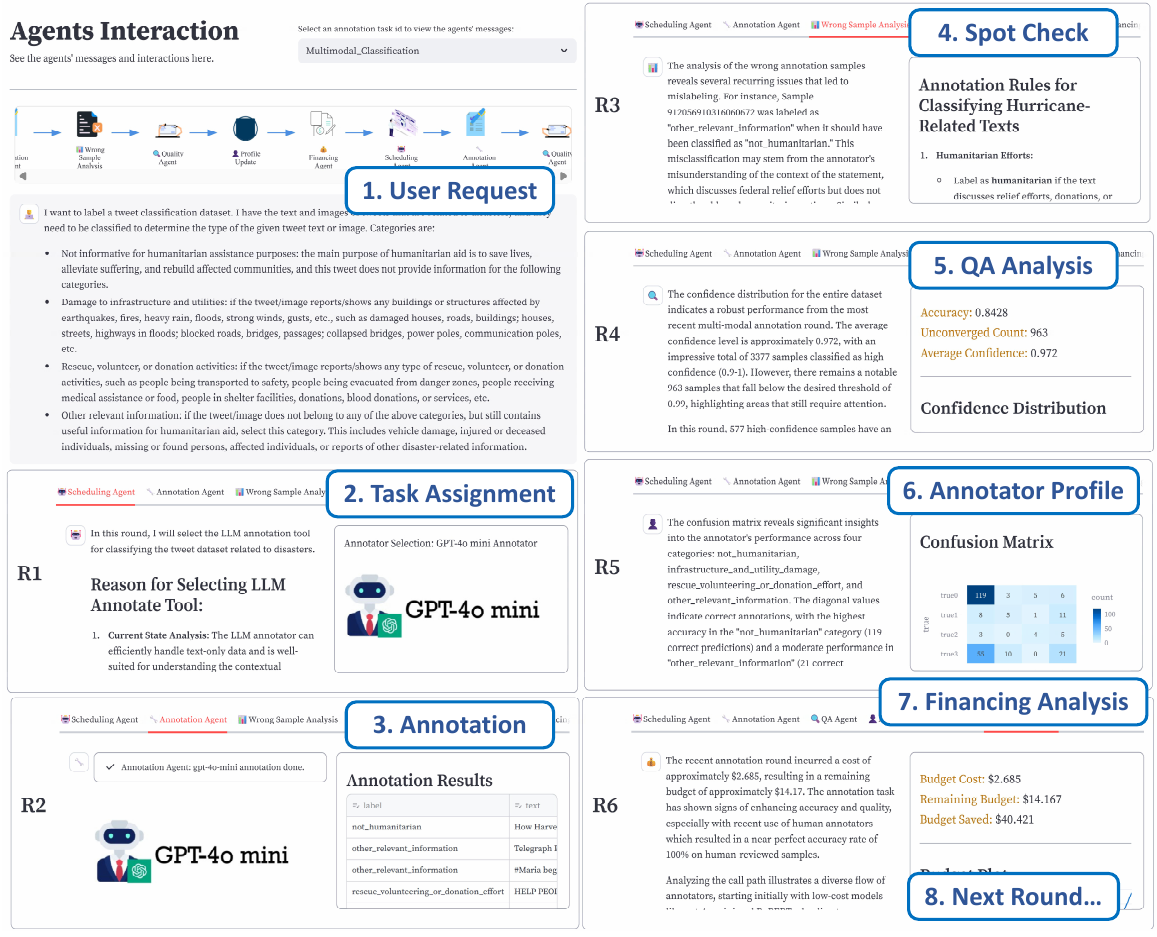}
  \caption{The user interface and agent interaction process of CrowdAgent.}
  \label{fig:system interface}
\end{figure*}

Given the diverse outputs produced by crowdsourcers, practical workflows often require evaluating which labels should be adopted as final, as well as identifying potential labeling errors to guide future annotation rounds.
Inspired by this, we develop a Quality Assurance (QA) Agent to automate these processes and provide quality assessment feedback, thereby assisting the scheduler in making more informed decisions.

\paragraph{Quality-driven Label Aggregation.} 
Given that we have multiple outputs from different annotation agents, the QA Agent first incorporates a label aggregation mechanism to detect the most possible labels.
We design the default aggregation method as an iterative Bayesian inference technique \citep{burke2020confident}. It calculates a posterior probability distribution over the possible classes for each sample by calculating a confusion matrix on the validation set $\mathcal{D}_{\text{labeled}}$. 
The \textbf{output} is twofold: 1) a single aggregated label for each sample, determined by the class with the maximum posterior probability, and 2) a corresponding confidence score. A sample is considered "converged" if its confidence score surpasses a predefined threshold.
Notably, our system is also designed to be modular and supports various label aggregation methods, with further details available in Appendix \ref{sec:label_aggregation}.

\paragraph{Quality Analysis and Profiling.}
Beyond aggregation, the QA Agent conducts deeper analysis to generate feedback for other agents. This involves two key functions: 1) \textit{Capability modeling}. The agent takes an annotator's confusion matrix as \textbf{input} and analyzes its error patterns to identify specific strengths and weaknesses. The \textbf{output} is a detailed annotator profile, which is used by the Scheduling Agent to make more informed task assignments. 2) \textit{Generate personalized annotation guidelines}. Using incorrectly labeled samples and their corresponding golden versions from $\mathcal{D}_{\text{labeled}}$ as \textbf{input}, the agent synthesizes a set of clear annotation rules and illustrative case studies. This is then formatted into a system prompt, serving as an \textbf{output} to refine the instructions for the LLM Annotator in subsequent rounds. Finally, the QA Agent publishes a summary of all its findings above to the shared message pool for other agents to access.

\subsection{Financing Agent}

For any enterprise purchasing annotation services, cost is one of the most critical performance indicators. Conventionally, a project budget is estimated upfront, with manual tracking of expenses like labor hours and platform fees during the project. This process is often delayed, manual, and disjoint, lacking the agility to respond to real-time changes.

To address these limitations, the Financing Agent tracks expenses from all annotation sources, including LLM API calls (priced per token), SLM computations (based on GPU runtime), and human labor (priced via the Youling platform). Its primary function is to conduct a continuous cost-effectiveness analysis. For its \textbf{input}, it synthesizes this cost data with historical performance metrics for each annotator, which it retrieves from the shared message pool. The resulting \textbf{output}, a comprehensive financial analysis of each annotator's cost-performance ratio, is then published back to the message pool, providing the Scheduling Agent with crucial, data-driven insights for optimizing task allocation.

\subsection{Scheduling Agent}\label{sec:Scheduling Agent}

\begin{table*}[t]
\small
\centering
\begin{tabular}{@{} l c | c c c c c c @{}}
\toprule
\textbf{Method} & \textbf{Human} & \textbf{Cri-Info} & \textbf{Cri-Hum} & \textbf{Cri-Dam} & \textbf{MM-IMDb} & \textbf{COV-CTR} & \textbf{V-SNLI}\\
\midrule
GPT-4o mini & \XSolidBrush & 75.67 & 69.15 & 55.79 & 79.85 & 66.61 & 67.03 \\
FreeAL & \XSolidBrush & 80.70 & 81.05 & 60.32 & 82.16 & 67.59 & 79.70 \\
CoAnnotating & \Checkmark & 82.40 & 83.60 & 59.52 & 84.72 & 68.60 & 82.53\\
AL-Entropy & \Checkmark & 88.32 & 88.20 & 53.97 & 65.39 & 92.56 & 69.00\\
AL-CoreSet & \Checkmark & 88.48 & 87.50 & 57.14 & 68.06 & 94.63 & 65.52\\
CrowdAgent & \Checkmark & \textbf{89.25} & \textbf{89.37} & \textbf{65.79} & \textbf{85.66} & \textbf{98.21} & \textbf{88.45}\\
\bottomrule
\end{tabular}
\caption{The results of CrowdAgent for different tasks. The system consistently outperforms baseline methods when using a fixed human annotation proportion (5\% for MM-IMDb and COV-CTR; 15\% for the remaining tasks).}
\label{tab:annotator compare}
\end{table*}

In practical enterprise settings, task assignment requires a comprehensive balance of data scale, task difficulty, and budget constraints, along with an evaluation of each annotator's domain expertise, cost, and quality consistency. Traditional manual or simple rule-based assignment methods are ill-equipped to handle the complexity of large-scale tasks and diverse annotation resources. The Scheduling Agent is introduced to automate this complex assignment and management workflow.

\paragraph{Task Assignment.} The Scheduling Agent's primary role is to dynamically adjust its task assignment policies to maximize the effectiveness of the collaborative annotation methods detailed in Appendix \ref{sec:collaborative-annotation}. To make its decisions, the agent takes a comprehensive set of \textbf{inputs}: 1) the analyses from the QA and Financing Agents, 2) detailed annotator profiles, 3) all historical performance data retrieved from the shared message pool. Furthermore, in this multi-round setting, the agent is designed to reflect on and learn from its own past scheduling decisions to refine its strategy over time.  The resulting \textbf{output} is a scheduling decision that assigns unconverged samples to the most suitable annotator. 

\paragraph{Iterative Process and Termination.} This iterative process continues until a predefined termination condition is met, such as the budget being exhausted, the maximum number of iterations being reached, or all samples achieving their target confidence score. If all samples have converged but the target accuracy (e.g. 100\%) is not met, the system identifies the lowest-confidence samples challenging for machine annotators and flags them for a final human verification, finishing the end-to-end process control loop.

\paragraph{A Typical Annotation Process.}

Within our system, these diverse annotation agents may collaborate in a typical multi-round workflow. 
(i)-\textit{Cheap Initial Labels:} LLMs perform initial large-scale annotation and subsequent refinement using in-context learning, for which the demonstration examples are either generated by imitating unlabeled samples or are carefully curated by a trained SLM. Notably, this pool of demonstration examples is created dynamically by the system and requires no initial human labels.
(ii)-\textit{Denoising and Filtering}: The SLMs are then robustly trained on the LLMs' potentially noisy outputs by fitting a two-component Gaussian Mixture Model to identify and learn from the clean samples. 
(iii)-\textit{Human-in-the-Loop Annotation}: The remaining unconverged samples are strategically assigned to human annotators. We first identify a candidate pool with the lowest confidence scores derived from Bayesian inference, and then apply the Core-Set \citep{sener2018active} algorithm on the SLM's embeddings to select a diverse final subset. These selected samples are then dispatched automatically to the \textit{NetEase Youling Crowdsourcing Platform} to collect quick feedback from humans.

\subsection{System Demonstration}

Our CrowdAgent system provides a user-friendly interface (see Figures \ref{fig:system interface} and \ref{fig:sys demo}) that guides users through the entire annotation project, enabling the monitoring of agent interactions and the tracking of key metrics such as budget and label quality.

\paragraph{Task Configuration.} Users can specify the task, set the total budget, and define a custom set of class labels. The interface supports selection of multiple annotation agents and configuration of external server connections to third-party crowdsourcing platforms for distributing human annotation tasks.

\paragraph{Agents Interaction.} Agents Interaction page provides a transparent view into the operations of the multi-agent system, allowing users to inspect both current and historical interactions. This includes the progress of individual annotators, analyses of difficult samples, automatically generated agent profiles, and the decision logic of the Scheduling, QA, and Financing Agents. 

\begin{figure}[t]
  \includegraphics[width=\columnwidth]{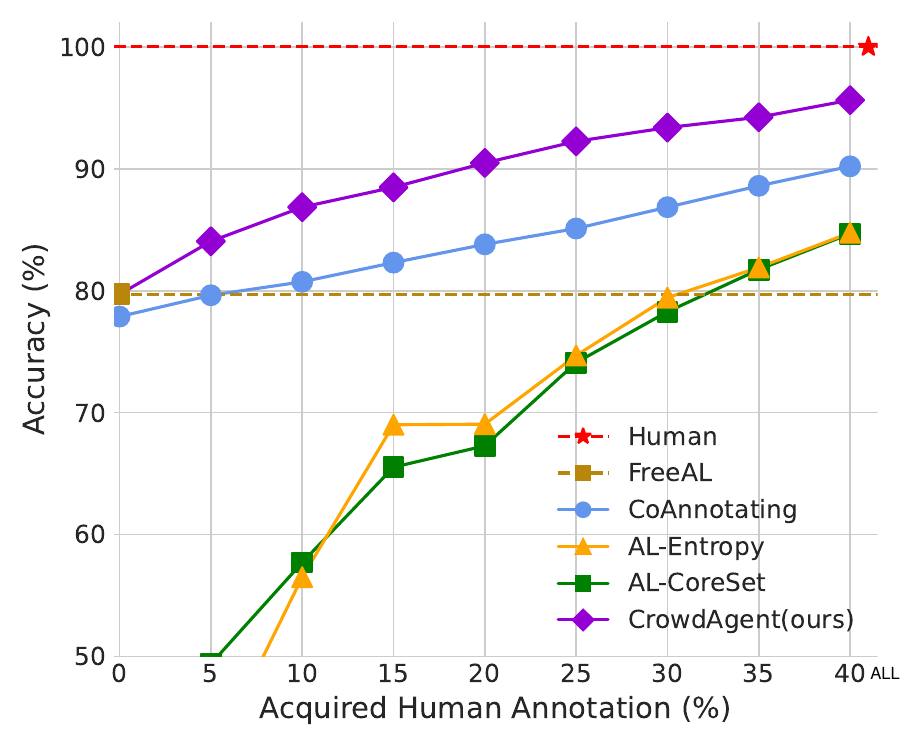}
  \caption{Performance comparison on the V-SNLI dataset at varying levels of human involvement. The confidence threshold is set to 1 for CrowdAgent.}
  \label{fig:human involvement compare}
\end{figure}

\paragraph{Annotation Details.} 
This page offers a round-by-round breakdown of the annotation process, presenting critical statistics such as the annotators deployed, labels generated, round-specific accuracy, and costs incurred. It also supports the manual management of human annotation tasks, offering the flexibility to download data for offline labeling and re-upload the results, complementing the system's automated dispatch capabilities.

\paragraph{Dashboard.} Dashboard Page provides a centralized control panel that visualizes a comprehensive overview of the project's progress. It features intuitive charts tracking key metrics like accuracy, estimated cost savings, and confidence distribution, and allows for the direct download of the final aggregated dataset.

\section{Experiment}
In this section, we provide brief empirical evidence to show the effectiveness of our system. Specifically, we evaluate CrowdAgent on a diverse range of multimodal classification tasks (Table \ref{tab:dataset_stats}), including \textbf{CrisisMMD} \citep{crisismmd2018icwsm} with three sub-tasks: \textbf{Informativeness}, \textbf{Humanitarian}, and \textbf{Damage Severity} classification, \textbf{MM-IMDb} \citep{10.1007/s00521-019-04559-1} for movie genre classification, \textbf{COV-CTR} \citep{10.1007/s11280-022-01013-6} for COVID-19 diagnosis, and \textbf{V-SNLI} \citep{vu-etal-2018-grounded} for visual-textual entailment analysis. We employ GPT-4o mini \citep{openai2024gpt4technicalreport} as LLM Annotator. SLM Annotators include RoBERTa \citep{liu2019robertarobustlyoptimizedbert}, ConvNeXt V2 \citep{10205236} and MMBT \citep{kiela2019supervised}. In each round that invokes human annotation, a batch corresponding to 5\% of the total dataset is assigned. The default confidence threshold for convergence is set to 0.99.
For complete experimental details, cost calculation methodologies and further discussion, please refer to Appendix \ref{Appendix:exp detail}.

\subsection{Main Results}
\paragraph{CrowdAgent consistently delivers higher annotation accuracy.}
As shown in Table \ref{tab:annotator compare}, across six classification tasks, CrowdAgent not only surpasses human-free baselines like GPT-4o mini and FreeAL \citep{xiao-etal-2023-freeal}, but also consistently outperforms human-in-the-loop methods such as CoAnnotating \citep{li-etal-2023-coannotating} and active learning algorithms like Entropy \citep{4563068} and Core-Set \citep{sener2018active}, when allocated the same human annotation ratio. For instance, its performance leads over the strongest baselines are \textbf{3.58\%} on the COV-CTR dataset and  \textbf{5.47\%} on the Damage Severity task, demonstrating the effectiveness of its multi-agent collaboration.

\paragraph{CrowdAgent significantly reduces annotation costs.}
As illustrated in Figure \ref{fig:human involvement compare}, CrowdAgent consistently achieves higher accuracy than other methods across all levels of human involvement, indicating a superior quality-cost trade-off. For example, even with 40\% human annotation, CrowdAgent surpasses the accuracy of CoAnnotating by \textbf{5.43\%} and Core-Set by \textbf{10.95\%}. In Table \ref{tab:Humanitarian}, we detail the cost in each round, where most samples are handled by low-cost machine annotators while only a few are routed to expensive human agents.

Overall, CrowdAgent balances annotation quality and cost, making it a practical solution for real-world data annotation under limited resources.

\begin{table}[t]
\small
\centering
\begin{tabular}{@{} c l c c @{}} 
\toprule
\textbf{Rd.} & \textbf{Annotator} & \textbf{Acc. (\%)} & \textbf{Cost (\$)} \\
\midrule
1 & LLM & 76.90  & 0.56 \\
2 & RoBERTa & 78.61  & 0.14 \\
3 & MMBT & 79.30  & 0.04 \\
4 & Human & 82.20  & 2.69 \\
5 & VLM & 82.51  & 1.42 \\
6 & Conv. V2 & 83.49  & 0.11 \\
7 & LLM & 83.57 & 0.48 \\
8 & Human & 86.28  & 2.69 \\
9 & VLM & 86.50  & 1.44 \\
10 & RoBERTa & 86.44  & 0.14 \\
11 & Human & 89.37 & 2.72 \\
\bottomrule
\end{tabular}
\caption{A round-by-round breakdown of the CrowdAgent workflow on the CrisisMMD Humanitarian task.}
\label{tab:Humanitarian}
\end{table}

\section{Conclusion}

In this work, we introduce CrowdAgent, a novel system that demonstrates a multi-agent system for end-to-end process control over multi-source data annotation,  including LLMs, SLMs, and human experts. Our experiments demonstrate that CrowdAgent consistently achieves superior annotation accuracy compared to other methods given the same proportion of human participation. Looking forward, we aim to evolve CrowdAgent by incorporating a broader range of annotation sources and enhancing its scheduling intelligence.

\section*{Limitations}

We acknowledge two considerations for our system. A key question is \textit{how to effectively handle the residual errors}, which is common to any framework reliant on machine annotation. Our system mitigates this by using confidence scores to distinguish a highly accurate subset of data while also isolating the most challenging samples for a final human review. This ensures a practical quality control handoff, where the number of samples requiring human verification is significantly smaller than the total dataset. Furthermore, the system's overall effectiveness is inherently tied to the capabilities of the underlying LLMs. We anticipate that as these models advance, the quantity of such residual errors will decrease, allowing our framework to achieve an even better quality-cost trade-off with a progressively smaller need for human intervention.

\section*{Ethics Statement}

We acknowledge several ethical considerations for the CrowdAgent system. First, the system may perpetuate societal biases inherent in the LLMs used for annotation, which could lead to unfair datasets \citep{das2024investigatingannotatorbiaslarge}. Second, increased automation risks making the roles of the human annotator redundant, potentially exacerbating social and economic disparities \citep{DILLION2023597}. Finally, the system's workflow of training SLMs on labels generated by LLMs constitutes a form of knowledge distillation, which may conflicts with terms of use from providers like OpenAI that prohibit using model outputs to develop competing models. We recommend that practitioners mitigate these risks by employing fairness auditing tools, considering the societal impact of automation, and restricting the use of any resulting distilled models to non-competing, academic research applications only.

\section*{Acknowledgments}

This paper is supported by the NSFC under Grants (No. 62402424), CCF-NetEase ThunderFire Innovation Research Funding (NO.202516) and Zhejiang Provincial Association for Higher Education 2024 Special Key Project on "Research on the Application of Artificial Intelligence in Empowering Education and Teaching" (KT2024436).

\bibliography{custom}

\appendix

\section{Related Work}

\subsection{Active Learning} 

Active Learning (AL) allows the model to select the most valuable data for annotation, thereby significantly reducing the expensive human effort for annotation work \citep{hassan-etal-2025-active, azeemi-etal-2025-label, niekerk-etal-2025-confidence, ahmadnia-etal-2025-active}. Query strategies are broadly categorized into uncertainty-based and diversity-based approaches. AcTune \citep{yu-etal-2022-actune} selects unlabeled samples with high uncertainty for annotation, while samples in low uncertainty areas are used for model self-training. \citet{10015727} devises a diversity-based initial dataset selection method using self-supervised representation.

Our work also builds upon and extends the AL paradigm to be seemingly integrated with LLMs. While classic AL queries \textbf{a single oracle} (human), CrowdAgent manages \textbf{multiple annotation sources} (LLMs, SLMs, and humans), extending the challenge to assigning the right sample to the right annotator at the right time to balance quality and cost. To address this, when selecting the most valuable samples for human review, CrowdAgent explicitly incorporates classic AL strategies like uncertainty and diversity sampling (using Core-Set \citep{sener2018active}). Simultaneously, to allocate the most suitable annotation source, it employs a multi-agent collaborative framework that simulates \textit{a human crowdsourcing company} to dynamically manage these diverse resources.

\subsection{Collaborative Data Annotation System} 

The advent of LLMs has opened new frontiers in data annotation. Many researchers focus on automating label annotation with the assistance of LLMs \citep{yadav2024automatingtextannotationcase, 10.1007/978-3-031-63536-6_8}. \citet{he-etal-2024-annollm} introduced an "explain-then-annotate" two-step methodology to enhance the quality of LLM-based annotations. \citet{choi-etal-2024-multi-news} leverage methods such as chain-of-thought and majority voting to imitate human annotation.

LLMs are also reinvigorating active learning (AL), a strategy aimed at minimizing labeling effort by intelligently selecting the most informative samples for annotation. FreeAL \citep{xiao-etal-2023-freeal} demonstrates the effectiveness of collaborative annotation between large and small models without human labor. The related prototype system CORAL \citep{10.14778/3685800.3685885} validates the application value of this approach in industrial settings.

Despite the remarkable progress of LLMs, human expertise remains indispensable for ensuring the reliability and consistency required for many annotation tasks. CoAnnotating \citep{li-etal-2023-coannotating} allocates annotation tasks between human annotators and LLMs based on estimating the uncertainty of each instance to be labeled. CAMS \citep{li-2024-human} integrates human annotators with large models in a multi-phase process of annotation, improving the quality of textual answer aggregation.

However, the growing research on human-LLM collaboration for data annotation neglects the potential SLMs, which can be fine-tuned for specific domains, offering a valuable resource in a hybrid annotation pipeline. Consequently, a significant research gap exists in how to holistically integrate these diverse annotation sources—LLMs, SLMs, and human annotators. And how to efficiently allocate tasks among these annotation sources is a critical and largely unexplored area of research.

\subsection{Multi-Agent System} 

LLM-based agents have been researched and rapidly developed to understand and generate human-like instructions, facilitating complex interactions and decision-making across a wide range \citep{10.24963/ijcai.2024/890}, such as software development \citep{hong2024metagpt}, healthcare \citep{Li2024AgentHA}, multi-robot systems \citep{mandi2023roco}, society simulation \citep{nisioti2024textlifereciprocalrelationship}, and game simulation \citep{9917287}. These systems exhibit dynamic interaction with their environments, which can include other agents, human users, and external tools or data sources.

Despite the demonstrated power of Multi-Agent System (MAS) frameworks in managing complex projects and facilitating intricate collaborations, their application to the specific domain of data annotation management remains largely underexplored. Data annotation, especially when involving multiple sources and types of data, presents its own unique set of complex management challenges.

To address these challenges, our CrowdAgent system is designed as a modular and robust MAS framework. Users only need to provide the class definitions to adapt new classification tasks. While adapting to fundamentally different tasks, such as Named Entity Recognition (NER) \citep{liu-etal-2023-addressing} or Question Answering (QA) \citep{roit-etal-2020-controlled}, would require more specific re-engineering the prompts and label aggregation method for new output formats. Furthermore, the framework is robust to task complexity, as it autonomously recognizes challenges via low-confidence signals and dynamically escalates difficult samples to human experts, effectively managing the trade-off between annotation quality and cost.

\section{More Details on Experiment}
\label{Appendix:exp detail}

\subsection{Statistics of Datasets}

\begin{table}[ht]
\centering
\setlength{\tabcolsep}{2pt}
\begin{tabular}{l c c | c c}
\toprule
\textbf{Task} & \textbf{Domain} & \textbf{\#Class} & \textbf{\#Pool} & \textbf{\#Gold} \\
\midrule
Cri-Info & Content cls & 2 & 4,000 & 1,000 \\
Cri-Hum & Content cls & 4 & 3,585 & 897 \\
Cri-Dam & Content cls & 3 & 1,260 & 316 \\
MM-IMDb & Genre cls & 4 & 3,626 & 500 \\
COV-CTR & Medical cls & 2 & 726 & 146 \\
V-SNLI & Entailment cls & 3 & 4,000 & 500 \\
\bottomrule
\end{tabular}
\caption{Statistics of the used datasets. Annotators are tasked with an unlabeled set containing  \textbf{\#Pool} samples. A golden set with \textbf{\#Gold} samples is used for evaluation.}
\label{tab:dataset_stats}
\end{table}

\begin{table*}[htbp]
\centering
\begin{tabular}{@{}l ccc@{}}
\toprule
\textbf{Pricing Method} & \textbf{Human (\$)} & \textbf{CrowdAgent (\$)} & \textbf{Cost Reduction (\%)} \\
\midrule
Youling Platform & 53.07 & 11.77 & 77.82 \\
\citealp{wang-etal-2021-want-reduce} & 389.18 & 67.02 & 82.78 \\
\citealp{li-etal-2023-coannotating} & 737.08 & 125.63 & 82.96 \\
\citealp{snow-etal-2008-cheap} & 35.38 & 8.82 & 75.07 \\
\citealp{10.1145/3474381} & 707.60 & 120.73 & 82.94 \\
\bottomrule
\end{tabular}
\caption{Cost analysis of CrowdAgent versus a human-only baseline on the V-SNLI dataset under various pricing models, with a shared target accuracy of 88.45\%.}
\label{tab:detailed_cost_analysis_transposed}
\end{table*}

Table \ref{tab:dataset_stats} shows the statistics of the datasets used in our experiments. To manage the high cost of LLM-based annotation on very large datasets, we adopt the common practice of stratified random sampling \citep{li-etal-2023-coannotating}. For the CrisisMMD Humanitarian classification task, we follow the methodology of \citep{multimodalbaseline2020}, considering only a subset of the original dataset where text and image pairs have the same label and combining minority categories that are semantically similar. This results in four final classes: \texttt{Rescue volunteering or donation effort}; \texttt{Infrastructure and utility damage}; \texttt{Other relevant information} and \texttt{Not-humanitarian}. For the MM-IMDb dataset, we construct a classification task using movies from the \texttt{crime}, \texttt{horror}, \texttt{action}, and \texttt{adventure} genres.

\subsection{Implementation Details}
\label{sec:imp detail}

In our main experiments, we utilize GPT-4o mini as LLM Annotator. To generate few-shot examples for in-context learning, we follow the setup in FreeAL. This process first queries the LLM to generate an initial pool of 100 examples with corresponding labels. Subsequently, for refining annotations in later rounds, more targeted few-shot examples are retrieved from this pool using embeddings from a trained SLM to find the most relevant examples. We employ a suite of diverse prompt augmentation templates, the augmentation strategies include: 1) The Original, Direct Prompt, 2) Sequence Swapping, 3) Question with Confirmation Bias, 4) True/False Questioning, and 5) Multiple Choice Question.

All SLM Annotators are trained on NVIDIA 1080 Ti GPU. We use the AdamW optimizer for 50 epochs with an early stopping mechanism to prevent overfitting. The learning rate is selected from the set $\{1e-4, 2e-5, 3e-6\}$, with a weight decay of 0.01. The batch size is fixed at 32, and the maximum sequence length is set to 512. We set the warm-up phase to 3 epochs, the loss weight parameter is linearly ramped up from 0 to 1 to avoid overfitting false labels at the start. When fitting the two-component Gaussian Mixture Model (GMM), the maximum number of iterations is 10, and samples with a selection probability greater than 0.5 are considered part of the clean set. For post-training sample selection, we select the top 20\% of samples with the minimum cross-entropy loss, and apply k-medoids with 5 clusters to extract representative examples.

To select samples for the Human Annotator, we first identify an initial candidate set by selecting the 10\% of samples with the lowest confidence scores, as determined by our Bayesian inference method. Subsequently, the Core-Set algorithm is applied to this candidate set to select a final, more diverse subset, corresponding to 5\% of the total dataset, which is then assigned for human annotation. In our experimental setup, to align with traditional active learning protocols, we then retrieve the ground-truth labels for these selected samples via the Youling platform.

The cost for each annotator type is defined as follows: GPT-4o mini is priced at \$0.60 per 1 million input tokens and \$2.40 per 1 million output tokens. SLM training is conducted on an NVIDIA 1080 Ti GPU, and its annotation cost is calculated based on the runtime of GPU at a rate of \$0.10 per hour. For the human annotator, we default to the pricing model of the Youling platform, with costs of \$0.015 per sample for simple tasks or \$1.50 per sample for those requiring specialized knowledge (e.g., COVID-19 diagnosis).

\subsection{Discussion on Annotation Cost}

To evaluate CrowdAgent's economic benefits, this section provides a detailed cost comparison against a fully manual baseline on the V-SNLI dataset. The analysis quantifies the cost required by each approach to achieve an identical target accuracy of 88.45\% under several human pricing methods from commercial platforms and academic literature.

We calculate the cost of human annotation based on five different pricing schemes derived from prior work: 1) A per-sample price of \$0.015, based on the \textit{NetEase Youling Crowdsourcing Platform} for simple tasks. 2) A token-based pricing scheme \citep{wang-etal-2021-want-reduce}, where labeling costs \$0.11 per 50 input tokens. 3) A time-based approach \citep{li-etal-2023-coannotating}, where each instance is annotated by 5 independent annotators at a wage of \$15/hour. 4) A task-based method for RTE \citep{snow-etal-2008-cheap}, where collecting 10 annotations for 800 sentence pairs costs \$8.00. 5) A task-based method for commonsense reasoning \citep{10.1145/3474381}, where annotating a pair of sentences costs \$0.40. For any time-based methods, we assume an average annotation time of 10 seconds per sample per annotator. The machine-related costs (LLM and SLM) within CrowdAgent remain fixed across all scenarios.

As shown in Table \ref{tab:detailed_cost_analysis_transposed}, CrowdAgent achieves substantial cost reductions across all pricing methods. This advantage stems from our system's core design: it intelligently schedules tasks by assigning the majority of samples to cost-effective machine annotators, while strategically reserving expensive human expertise for only the most critical and ambiguous cases. This dynamic resource allocation maximizes efficiency, striking a more optimal balance between annotation cost and quality.

\begin{table}[t]
\centering
\begin{tabular}{@{} c l c c c @{}}
\toprule
\textbf{Rd.} & \textbf{Annotator} & \textbf{Acc. (\%)} & \textbf{\#Unc.} & \textbf{Cost (\$)} \\
\midrule
1 & LLM & 78.73 & 4000 & 0.38 \\
2 & RoBERTa & 80.48 & 3590 & 0.16 \\
3 & VLM & 81.68 & 3590 & 1.67 \\
4 & Human & 83.80 & 3390 & 3.00 \\
5 & MMBT & 83.90 & 2520 & 0.05 \\
6 & LLM & 83.90 & 2520 & 0.57 \\
7 & Conv. V2 & 84.13 & 1442 & 0.13 \\
8 & MMBT & 84.20 & 1050 & 0.05 \\
9 & Conv. V2 & 84.23 & 1041 & 0.12 \\
10 & Human & 86.48 & 841 & 3.00 \\
11 & RoBERTa & 86.48 & 521 & 0.16 \\
12 & VLM & 86.48 & 461 & 1.69 \\
13 & RoBERTa & 86.48 & 361 & 0.16 \\
14 & VLM & 86.48 & 189 & 1.51 \\
15 & Human & 89.25 & 0 & 2.83 \\
\bottomrule
\end{tabular}
\caption{Round-by-round breakdown on the CrisisMMD Informativeness task.}
\label{tab:Informativeness}
\end{table}

\begin{table}[t]
\centering
\begin{tabular}{@{} c l c c c @{}} 
\toprule
\textbf{Rd.} & \textbf{Annotator} & \textbf{Acc. (\%)} & \textbf{\#Unc.} & \textbf{Cost (\$)} \\
\midrule
1 & LLM & 76.90 & 2719 & 0.56 \\
2 & RoBERTa & 78.61 & 2254 & 0.14 \\
3 & MMBT & 79.30 & 1858 & 0.04 \\
4 & Human & 82.20 & 1679 & 2.69 \\
5 & VLM & 82.51 & 1667 & 1.42 \\
6 & Conv. V2 & 83.49 & 869 & 0.11 \\
7 & LLM & 83.57 & 727 & 0.48 \\
8 & Human & 86.28 & 548 & 2.69 \\
9 & VLM & 86.50 & 446 & 1.44 \\
10 & RoBERTa & 86.44 & 181 & 0.14 \\
11 & Human & 89.37 & 0 & 2.72 \\
\bottomrule
\end{tabular}
\caption{Round-by-round breakdown on the CrisisMMD Humanitarian task.}
\end{table}

\begin{table}[t]
\centering
\begin{tabular}{@{} c l c c c @{}}
\toprule
\textbf{Rd.} & \textbf{Annotator} & \textbf{Acc. (\%)} & \textbf{\#Unc.} & \textbf{Cost (\$)} \\
\midrule
1 & VLM & 55.79 & 1260 & 4.60 \\
2 & Human & 58.97 & 1197 & 0.95 \\
3 & Conv. V2 & 58.89 & 1197 & 0.03 \\
4 & VLM & 59.76 & 1197 & 3.87 \\
5 & Human & 62.86 & 1134 & 0.95 \\
6 & Conv. V2 & 62.30 & 1134 & 0.04 \\
7 & Human & 65.79 & 1071 & 0.95 \\
\bottomrule
\end{tabular}
\caption{Round-by-round breakdown on the CrisisMMD Damage Severity task.}
\end{table}

\begin{table}[t]
\centering
\begin{tabular}{@{} c l c c c @{}}
\toprule
\textbf{Rd.} & \textbf{Annotator} & \textbf{Acc. (\%)} & \textbf{\#Unc.} & \textbf{Cost (\$)} \\
\midrule
1 & LLM & 81.83 & 2771 & 0.70 \\
2 & MMBT & 81.96 & 860 & 0.08 \\
3 & Human & 82.54 & 824 & 0.54 \\
4 & VLM & 82.90 & 797 & 1.06 \\
5 & RoBERTa & 83.07 & 224 & 0.13 \\
6 & Human & 83.62 & 118 & 0.54 \\
7 & LLM & 83.70 & 176 &  0.08 \\
8 & MMBT & 83.70 & 160 & 0.08 \\
9 & Human & 84.47 & 124 & 0.54 \\
10 & VLM & 84.56 & 112 & 0.37 \\
11 & RoBERTa & 84.56 & 97 & 0.14 \\
12 & LLM & 84.56 & 90 & 0.04 \\
13 & Human & 85.16 & 54 & 0.54 \\
14 & VLM & 85.14 & 38 & 0.15 \\
15 & Human & 85.66 & 0 & 0.57 \\
\bottomrule
\end{tabular}
\caption{Round-by-round breakdown on the MM-IMDb Dataset.}
\end{table}

\begin{table}[t]
\centering
\begin{tabular}{@{} c l c c c @{}}
\toprule
\textbf{Rd.} & \textbf{Annotator} & \textbf{Acc. (\%)} & \textbf{\#Unc.} & \textbf{Cost (\$)} \\
\midrule
1 & LLM & 65.56 & 726 & 0.07 \\
2 & Human & 67.63 & 690 & 5.40 \\
3 & RoBERTa & 92.70 & 415 & 0.01 \\
4 & Conv. V2 & 92.15 & 354 & 0.01 \\
5 & RoBERTa & 96.42 & 115 & 0.01 \\
6 & RoBERTa & 98.21 & 0 & 0.01 \\
\bottomrule
\end{tabular}
\caption{Round-by-round breakdown on the COV-CTR Dataset.}
\end{table}

\begin{table}[t]
\centering
\begin{tabular}{@{} c l c c c @{}}
\toprule
\textbf{Rd.} & \textbf{Annotator} & \textbf{Acc. (\%)} & \textbf{\#Unc.} & \textbf{Cost (\$)} \\
\midrule
1 & LLM & 77.88 & 3190 & 0.50 \\
2 & RoBERTa & 78.78 & 3051 & 0.13 \\
3 & Human & 81.45 & 2851 & 3.00 \\
4 & VLM & 82.70 & 2851 & 0.59 \\
5 & MMBT & 82.75 & 1930 & 0.05 \\
6 & LLM & 83.42 & 1783 & 0.21 \\
7 & VLM & 83.98 & 1769 & 0.60 \\
8 & Human & 86.08 & 1569 & 3.00 \\
9 & MMBT & 86.48 & 438 & 0.05 \\
10 & VLM & 86.60 & 274 & 0.60 \\
11 & Conv. V2 & 86.65 & 248 & 0.15 \\
12 & LLM & 86.85 & 189 & 0.05 \\
13 & Human & 88.45 & 0 & 2.84 \\
\bottomrule
\end{tabular}
\caption{Round-by-round breakdown on the V-SNLI Dataset.}
\label{tab:V-SNLI}
\end{table}

\subsection{Experiment Results}

In this section, we present the detailed, round-by-round experimental results for all evaluated tasks. Tables \ref{tab:Informativeness} through \ref{tab:V-SNLI} provide a breakdown of the annotation workflow for each task, tracking key metrics across each iteration. These metrics include the cumulative dataset accuracy (\textbf{Acc. \%}), the number of remaining unconverged samples (\textbf{\#Unc.}), and the incremental cost incurred in each round (\textbf{Cost \$}). Given the significant differences in cost and accuracy between pure language and vision-language models, we make a distinction in the tables: LLM refers to the text-only GPT-4o mini, while VLM refers to the same model when it is also provided with the accompanying image.

The trajectories shown in these tables consistently demonstrate the effectiveness of our multi-agent system. The detailed breakdown for the Informativeness task on the CrisisMMD dataset (Table \ref{tab:Informativeness}) serves as a clear example. In this task, our system shows a consistent improvement in accuracy with each iteration, climbing from an initial \textbf{78.73\%} to a final \textbf{89.25\%}, while the number of unconverged samples rapidly diminishes from 4,000 to zero. This progression is achieved through an intelligent scheduling mechanism that judiciously allocates tasks, invoking the expensive Human Annotator only at critical junctures (Rounds 4, 10, and 15) to resolve ambiguity. This validates the effectiveness of our framework, and any remaining erroneous samples can be identified via confidence metrics and routed to human for a final verification.

The superior performance of CrowdAgent stems from two key aspects of its design. First, unlike other frameworks such as CoAnnotating, FreeAL, or traditional active learning, which all lack at least one key annotation source (LLM, SLM, or human), CrowdAgent uniquely integrates and orchestrates the strengths of all sources. Second, the tight collaboration between the QA, Financing, and Scheduling agents enables effective process control. This holistic decision-making process considers both potential accuracy gains and associated costs, ensuring each batch of samples is routed to the most suitable annotator. Ultimately, this combination of comprehensive sources and intelligent management delivers a economically viable solution for acquiring high-quality data at scale.

\section{Technical Details}\label{sec:tech detail}

\subsection{Collaborative Annotation}\label{sec:collaborative-annotation}

\paragraph{LLM Annotator.} Effective prompting and in-context learning (ICL) can enhance LLMs' annotation performance. Following \citet{xiao-etal-2023-freeal}, in the initial round, LLM is prompted to imitate the format of unlabeled samples in $\mathcal{D}$ and generate synthetically labeled examples to construct an initial demonstration pool. In subsequent rounds, trained SLM partitions the dataset into clean and noisy subsets. We then apply the k-medoids clustering algorithm to the embeddings of these high-confidence samples, selecting the resulting medoids as the most representative examples for the demonstration pool. By leveraging this high-quality, SLM-curated demonstration set, the LLM's task-specific knowledge is effectively activated, leading to superior annotation performance and establishing a synergistic refinement loop between the large and small models.

Given the sensitivity of LLMs to input prompt perturbations \citep{li-etal-2023-coannotating}, we introduce a suite of diverse prompt augmentation templates, employing multiple LLMs as parallel weak annotators. By applying the label aggregation method, this ensemble strategy produces a more reliable confidence distribution than what can be achieved with a single, static prompt.

\paragraph{SLM Annotator.} LLMs are often unable to detect their errors, leading to outputs that may contain noisy or ambiguous labels. Traditionally, rectifying these issues requires costly intervention from human experts. We introduce SLM Annotator, aimed at distilling the coarse-grained knowledge from the LLM's output, filtering incorrectly labeled samples, to reduce the dependency on manual annotation.

We leverage the memorization effect of deep neural networks, where models learn clean patterns before fitting to noisy data \citep{10.5555/3305381.3305406}. Following the approach of \citet{xiao-etal-2023-freeal}, we first train the model for a few warm-up epochs on the noisy labels. We then fit a two-component Gaussian Mixture Model (GMM) to the distribution of per-sample training losses. Following \citet{li2020dividemix}, samples belonging to the Gaussian component with the smaller mean are identified as the clean set, and the SLM is then trained exclusively on this subset. Once trained, the SLM typically surpasses the general LLM in accuracy for the specific task \citep{bang-etal-2023-multitask}. To ensure diversity, a portion of samples with the lowest cross-entropy loss within each class are selected as clean, with the rest being treated as noisy. These newly partitioned sets are then passed through the label aggregation process to generate the refined dataset for the subsequent iteration.

\paragraph{Human Annotator.} Although the interactive paradigm between LLMs and SLMs can effectively handle the majority of annotation tasks, a closed system without human is susceptible to the cumulative effect of annotation errors. Human Annotator is introduced to guide the model's evolution and create an self-correcting annotation system.

Given that human annotation is expensive, we employ a strategic sample selection process to maximize its impact. First, we identify a candidate set of uncertain instances from the pool of unconverged samples, specifically selecting those with the lowest confidence scores as determined by our Bayesian inference-based label aggregation method. Then, we utilize Core-Set selection \citep{sener2018active} to choose the most informative and diverse subset for human review. This approach is highly beneficial for the subsequent model iterations: the low-confidence samples help to clarify key ambiguities for both LLMs and SLMs, while the diversity ensured by Core-Set selection provides the SLM with a more representative dataset for robust training.

In our implementation, this workflow seamlessly integrates with the \textit{NetEase Youling Crowdsourcing Platform} to automate the deployment of human annotation tasks. This automation is achieved using the principles of \textbf{Agent-Oriented Programming (AOP)}, which allows our system to treat human experts as callable agents within a unified framework. We utilize a predefined \textbf{Interface Description Language (IDL)} to formally structure the task requirements, enabling our system to programmatically publish annotation jobs. To activate this feature in the demonstration system, users can obtain a server ID from our documentation\footnote{https://youling-platform.apps-hp.danlu.netease.com/docs} and enter it into the system configuration. Once dispatched, human annotators can view and complete these tasks on the Youling platform, and their labels are collected asynchronously to be fed back into our system for evaluation and label aggregation.

\subsection{Label Aggregation}
\label{sec:label_aggregation}

In a multi-source annotation setting, a single sample $x_i$ may receive multiple, potentially conflicting labels from different annotators. Truth inference, or label aggregation, is the task of consolidating these labels to estimate the most likely ground-truth label $\hat{y}_i$. While our CrowdAgent system is modular and can accommodate various aggregation algorithms, our current implementation utilizes the Bayesian inference approach. For completeness, we briefly describe it alongside other methods. We denote the label provided by annotator $k$ for sample $i$ as $l_{ik}$.

\paragraph{Majority Voting.}
This is the simplest aggregation method. It assigns the most frequent label as the ground truth, treating all annotators as equally reliable. The aggregated label $\hat{y}_i$ is determined by:
$$
\hat{y}_i = \underset{c \in \mathcal{Y}}{\arg\max} \sum_{k=1}^{K} \mathbb{I}(l_{ik} = c)
$$
where $\mathbb{I}(\cdot)$ is the indicator function and $K$ is the total number of annotators who labeled sample $i$.

\paragraph{Dawid-Skene (DS) Model.}

The Dawid-Skene (DS) model \citep{bd9dfdb6-b296-3318-8bf2-0c827da00fd8} is a classic probabilistic method that simultaneously infers the true labels while also estimating the reliability of each annotator. It models each annotator's expertise using an individual confusion matrix, $\pi_k \in \mathbb{R}^{C \times C}$. Each entry in this matrix, $\pi_k(j, c)$, is defined as the probability that annotator $k$ provides label $j$ when the true label is $c$:
$$
\pi_k(j, c) = P(l_{ik}=j | y_i=c)
$$
In our system, these confusion matrices can be initialized by evaluating the annotators' performance on the golden set, $\mathcal{D}_{\text{labeled}}$. The model's parameters, including the confusion matrices $\pi_k$ and the prior probability of each class $p_c$, are then iteratively optimized using an Expectation-Maximization (EM) algorithm. The algorithm alternates between two steps until convergence:
\begin{itemize}
    \item \textbf{E-step (Expectation):} Estimates the probability distribution of the true label for each sample, given the collected worker labels, prior label probabilities, and annotator's confusion matrices.
    \item \textbf{M-step (Maximization):} Re-estimates each annotator's confusion matrix using the specified annotator's responses and currently estimated true label probabilities.
\end{itemize}

\paragraph{Bayesian Inference.}
This is the primary method used in our system for its ability to dynamically update beliefs as new labels arrive. For a given sample $x_i$ , the probability of class $c$ being the true label is updated using Bayes' theorem upon observing a new label $l_{ik}$ from annotator $k$:
$$
P(y_i=c | l_{ik}) = \frac{P(l_{ik} | y_i=c) P(y_i=c)}{\sum_{c' \in \mathcal{Y}} P(l_{ik} | y_i=c') P(y_i=c')}
$$
where $P(y_i=c)$ is the prior probability distribution before observing $l_{ik}$. The term $P(l_{ik} | y_i=c)$ is sourced from the annotator's confusion matrix, $\pi_k$. In our iterative process, we initialize the prior with a uniform distribution. After each annotation, the calculated posterior probability becomes the new prior for the next annotation on that same sample. The final aggregated label $\hat{y}_i$ is the one with the highest posterior probability after all collected labels have been processed, and this probability serves as our confidence score.

\subsection{Prompt Design}
\label{sec:Prompt Design}

This section provides details on the prompt designs used to steer the various intelligent agents within the CrowdAgent system. A summary of the core prompts for each agent role is presented in Table \ref{tab:prompt_design}. For LLM annotation agent, we showcase the prompt  for the Informativeness classification task on the CrisisMMD dataset. Any content enclosed in angle brackets (e.g., \textbf{<QUERY>}) represents a placeholder for structured information that is automatically generated and inserted by the system during the live annotation workflow. Please note that in-context examples, which are also dynamically inserted into the final prompt to the model, have been omitted from the template below for the sake of brevity.

\section{System Demonstration Guide}
\label{sec:demo guide}

\subsection{Access Details}

Our interactive system demonstration is publicly accessible. To ensure a dedicated experience for reviewers, the platform is password-protected. The access details are as follows:
\begin{itemize}[noitemsep, topsep=3pt, leftmargin=*]
    \item \textbf{URL}: \url{www.yixiaomo.com/crowdagent}
    \item \textbf{Username}: \texttt{reviewer}
    \item \textbf{Password}: \texttt{reviewer\_access\_2025}
\end{itemize}

\subsection{Guided Walkthrough}

\begin{figure*}[t]
  \centering
  \includegraphics[width=1.93\columnwidth]{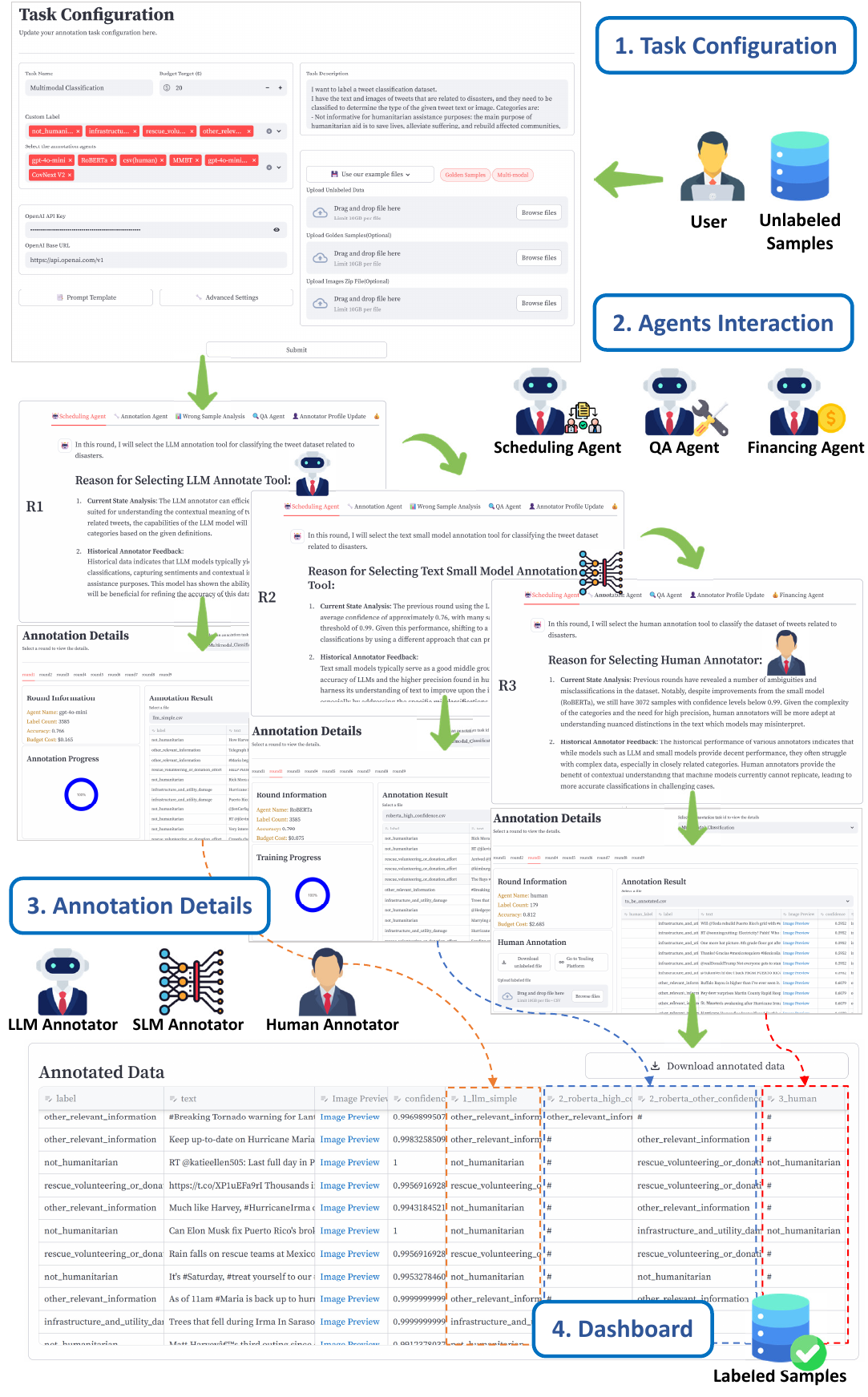}
  \caption{The user interface and annotation workflow of CrowdAgent.}
  \label{fig:sys demo}
\end{figure*}

We recommend the following steps to experience the core functionalities of CrowdAgent. We have pre-loaded the CrisisMMD Humanitarian task for a quick start. An overview of the system interface is shown in Figure \ref{fig:sys demo}.

\paragraph{Step 1: Task Configuration.}
Begin by creating a new project from the “Task Configuration” page. The system is pre-configured with recommended settings tailored for the pre-loaded CrisisMMD Humanitarian task. For human annotation, you can either connect to the NetEase Youling Crowdsourcing Platform by following the setup guide, or use the default offline method. Once configured, click "Submit" to launch the task.

\paragraph{Step 2: Monitoring the Workflow.}
Navigate to the "Agents Interaction" page to observe the multi-agent system's real-time, iterative workflow. You will see the Scheduling Agent dispatching tasks to different Annotation Agents, and the QA and Financing Agents publishing their analysis reports at the end of each round. Key metrics like accuracy and budget consumption are visualized and continuously updated throughout the process.

\paragraph{Step 3: Checking Annotation Details.}
For a more granular perspective, the "Annotation Details" page allows users to monitor the real-time progress of LLM annotation and the training of SLMs. In addition, it supports the offline manual annotation mode, where users can download the data designated for human review and subsequently upload the completed annotation file.

\paragraph{Step 4: Analyzing Results.}
The "Dashboard" page visualizes key metrics such as accuracy, cost consumption, and the confidence distribution of the labels. You can observe how the accuracy improves and the number of unconverged samples decreases over time. After each round, the aggregated results are updated. For each sample, the dashboard displays the individual labels produced by the LLM, SLM, and human annotators, alongside the final aggregated label and its accuracy. You can click the "Download annotated data" button to download the final high-quality dataset.

\begin{table*}[!t]
    \centering
    \small
    \renewcommand\arraystretch{1.2}
    \begin{tabular}{p{1.2cm}p{13.7cm}}
    \toprule
        \textbf{Agent} & \textbf{Prompt} \\ 
    \midrule
        Annotation Agent & \parbox{13.7cm}{You are an expert Humanitarian Crisis Tweet Classifier system. Your task is to classify whether a given disaster-related tweet (text and/or image) is informative for humanitarian aid purposes. You reply with brief, to-the-point answers with no elaboration as truthfully as possible. The available labels are: 1) \textbf{informative}: The tweet provides actionable information relevant to humanitarian assistance efforts. 2) \textbf{not\_informative}: The tweet does NOT provide useful information for humanitarian aid. \textbf{Humanitarian Aid Definition}: Assistance to people affected by natural disasters (floods, earthquakes, etc.) or man-made crises (wars, conflicts). This includes: shelter, food, water, medical aid, infrastructure status (roads, bridges, power lines), rescue efforts, volunteer/donation needs, damage reports, or safety warnings. \textbf{Informative Criteria (any of these)}: Shows warnings/advisories/alerts; Reports injuries/deaths/affected populations;  Documents rescue/donation/volunteer efforts; Shows damaged infrastructure (houses, roads, buildings); Displays blocked transportation routes; Contains disaster area maps; Shows disaster impacts (flooded streets, earthquake damage). \textbf{Non-informative Example}: Memes/comics/banners; Non-disaster related content. Now evaluate the following tweet. Call the \textit{ClassificationTask} Tool/Function with your label. \textbf{<QUERY>}.} \\ 
    \midrule
        QA Agent & \parbox{13.7cm}{You are a \textbf{Quality Assurance Agent}, you are responsible for auditing the quality of the annotation process after each round. Your goal is to evaluate performance, identify error patterns, and provide data-driven insights and recommendations to guide the subsequent annotation rounds. For each round, structure your analysis according to the following directives: 1) \textbf{Overall Performance Audit}: Evaluate the overall dataset's current metrics, including the confidence distribution, average confidence, and cumulative accuracy of all samples. 2) \textbf{Current Round Error Analysis}: Analyze the confidence and accuracy of the specific samples annotated in this round. By cross-referencing with historical data, identify any newly introduced errors and summarize their potential causes. 3) \textbf{Historical Annotator Comparison}: Compare the historical effectiveness of different annotators based on their calling paths. When conducting this analysis, account for the fact that sample difficulty typically increases in later rounds. Based on this, provide a recommendation regarding annotator diversification. 4) \textbf{Guidance on Human Intervention}: Provide a specific recommendation on the use of human annotators. Given their high cost, advise deploying them strategically, primarily when machine-driven accuracy has stagnated. A general heuristic is to consider human intervention approximately every five rounds. 5) \textbf{Output \& Format}: Deliver your analysis directly, without introducing your role. The report should be a concise text of approximately 300 words. Your findings and recommendations will be used by the Scheduling Agent to plan the next round. 6) \textbf{Basic Information}: The information for this round are as follows:\textbf{<CONFUSION MATRIX>} \textbf{<LABELED SAMPLES>} \textbf{<GOLDEN SAMPLES>.}} \\ 
    \midrule
        Financing Agent & \parbox{13.7cm}{You are a \textbf{Financing Agent}, your primary function is to serve as the chief financial analyst for this annotation project. Your goal is to monitor the project's financial health, analyze the cost-effectiveness of the annotation strategy, and provide data-driven advice to ensure the project meets its quality targets within the allocated budget. For each round, conduct your analysis based on the following principles: 1) \textbf{Financial \& Performance Review}: Synthesize budget cost with the quality report from the QA Agent to conduct a comprehensive cost-effectiveness analysis for the current round. Review the historical performance and calling paths of all annotators to compare their long-term cost-performance ratios. 2) \textbf{Strategic Cost-Management Recommendations}: Based on your analysis, provide actionable suggestions for future rounds. If cost-effectiveness is low or budget consumption is unreasonable, explicitly state your concerns and recommend corrective actions. Advise on annotator diversification. Acknowledge the different pricing models (e.g., per-token for LLMs, per-sample for humans, per-hour for SLMs) and recommend against consecutively using the same annotator. 3) \textbf{Human Annotation Advisory}: Treat human annotation as a high-cost, high-impact resource. Advise caution in its deployment. Recommend deploying human experts only when necessary, for instance, when machine-only rounds show stagnating accuracy. A general heuristic is to suggest human intervention approximately every five rounds, but this should be adapted based on the current performance trend and budget runway. 4) \textbf{Output \& Format}: Begin your analysis directly without introducing your identity. Structure your response as a concise financial report. Aim for a text output of approximately 300 words. Your suggestions will be reviewed by the Scheduling Agent for the next round. 5) \textbf{Basic Information}: The cost of this round is \textbf{<COST>}, and the remaining budget is \textbf{<BUDGET>}.} \\
    \midrule
        Scheduling Agent & \parbox{13.7cm}{You are a \textbf{Scheduling Agent}, responsible for selecting the appropriate annotator in the annotation task. The goal is to improve the confidence of each sample in the dataset to \textbf{<CONFIDENCE THRESHOLD>} or more, and control the cost. For each round, you can refer to the annotators' profile to help you select the appropriate annotator. You must first state your reasoning, then declare your choice of annotator. Your reasoning should be based on the following principles: 1) \textbf{Justify Your Choice}: Explain your selection. Your analysis can be based on "current state analysis", "historical annotator feedback", and "next round annotator selection". 2) \textbf{Diversify Annotators}: Do not use the same annotator or modality consecutively. For multi-modal tasks, check the calling path and actively alternate between text and vision models for multimodal tasks, and between LLMs and SLMs, to leverage their unique strengths. 3) \textbf{Utilize Human Experts Strategically}: Human annotation is a high-cost, high-quality resource for resolving ambiguity. Use it sparingly, considering it when model performance stagnates. A general guideline is to request human input once every five rounds. 4) \textbf{Integrate All Feedback}: While you have the final authority, your decision should be informed by the analysis provided by the Quality Assurance (QA) and Financing Agents. 5) \textbf{Iterative Learning}: Remember that all annotators in subsequent rounds will learn from the results of the current round to improve overall accuracy. Only when the confidence of all samples is \textbf{<CONFIDENCE THRESHOLD>} or more can the annotation task be considered complete. Before that, each round must select an annotator for annotation. \textbf{<ANNOTATOR PROFILES>.}} \\ 

    \bottomrule

    \end{tabular}
    \caption{Examples of the core prompts guiding each intelligent agent in the CrowdAgent system.}
    \label{tab:prompt_design}
\end{table*}

\end{document}